\documentclass{article}

\usepackage{arxiv}

\usepackage[utf8]{inputenc} 
\usepackage[T1]{fontenc}    
\usepackage{hyperref}       
\usepackage{url}            
\usepackage{booktabs}       
\usepackage{amsfonts}       
\usepackage{nicefrac}       
\usepackage{microtype}      
\usepackage{lipsum}		
\usepackage{graphicx}
\usepackage{natbib}
\usepackage{doi}

\usepackage{amsmath,amssymb}
\usepackage{algorithm}
\usepackage{color,colortbl}
\usepackage[noend]{algpseudocode}

\hypersetup{
    colorlinks=false,
    pdfborder={0 0 0},
    pdfborderstyle={/S/U/W 0},
}

\usepackage[table]{xcolor}
\usepackage{adjustbox}
\usepackage{caption,subcaption}
\usepackage{tikz}
\usetikzlibrary{shapes.arrows}
\usepackage{tikz-cd}

\tikzset{
    myarrow/.style={
        draw,
        single arrow,
        minimum height=9ex,
        single arrow head extend=1ex
    }
}

\newcommand{\arrowdown}{%
\tikz [baseline=-1ex]{\node [myarrow,rotate=-90] {};}
}

\definecolor{Gray}{gray}{0.9}

\title{The Privacy Issue of Counterfactual Explanations: Explanation Linkage Attacks}

\author{ \href{https://orcid.org/0000-0003-3784-826X}{\includegraphics[scale=0.06]{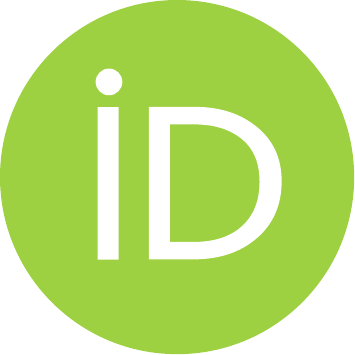}\hspace{1mm}Sofie Goethals}
\\
	Department of Engineering Management\\
	University of Antwerp\\
	Antwerp, Belgium\\
	\texttt{sofie.goethals@uantwerpen.be} \\
	\And 
	Kenneth Sörensen \\
	Department of Engineering Management\\
	University of Antwerp, Belgium\\
\And 
	David Martens \\
	Department of Engineering Management\\
	University of Antwerp, Belgium\\
}

\hypersetup{
pdftitle={The Privacy Issue of Counterfactual Explanations: Explanation Linkage Attacks},
}

\begin{document}
\maketitle

\begin{abstract}
Black-box machine learning models are being used in more and more high-stakes domains, which creates a growing need for Explainable AI (XAI). Unfortunately, the use of XAI in machine learning introduces new privacy risks, which currently remain largely unnoticed. We introduce the  \emph{explanation linkage attack}, which can occur when deploying instance-based strategies to find counterfactual explanations. 
To counter such an attack, we propose $k$-anonymous counterfactual explanations and introduce \emph{pureness} as a new metric to evaluate the \emph{validity} of these $k$-anonymous counterfactual explanations. Our results show that making the explanations, rather than the whole dataset, $k$-anonymous, is beneficial for the quality of the explanations. 
\end{abstract}


\section{Introduction}
Black-box models are used for decisions in more and more high-stakes domains such as finance, healthcare and justice, increasing the need to explain these decisions and to make sure that they are aligned with how we want the decisions to be made~\cite{molnar2020interpretable}. As a result, the interest in interpretability techniques for machine learning and the development of various techniques has soared~\citep{molnar2020interpretable}. At the moment, however, there is no consensus on which technique is best for which specific use case. Within the field of Explainable AI (XAI), we focus on a popular local explanation technique: counterfactual explanations~\citep{martens2014explaining,wachter2017counterfactual}.

\emph{Counterfactual explanations}, which are used to explain predictions of individual instances, are defined as the smallest change to the feature values of an instance that alters its prediction~\citep{martens2014explaining,molnar2020interpretable}. \emph{Factual instances} are the original instances that are explained and the \emph{counterfactual instance} is the original instance with the updated values from the explanation. An example of a \emph{factual instance}, \emph{counterfactual instance} and \emph{counterfactual explanation} for a credit scoring context can be seen in Figure~\ref{cf_explanation}. \emph{Lisa} is the \emph{factual instance} here, whose credit gets rejected. \emph{Fiona}, a nearby instance in the training set whose credit was accepted, is selected as \emph{counterfactual instance} by the algorithm and based on \emph{Fiona},  \emph{Lisa} receives a \emph{counterfactual explanation} that states which features to change to receive a positive credit decision. 
These explanations can serve multiple objectives: they can be used for model debugging by data scientists or model experts, to justify decisions to end users or provide actionable recourse, to detect bias in the model, to increase social acceptance, to comply with GDPR, etc.~\citep{aivodji2020model,molnar2020interpretable}.


\begin{table}[h]
\centering
\begin{tabular}{|lllllll|}
\hline
\multicolumn{7}{|l|}{\textbf{Factual instance}}                                                                                                                                                                                                                              \\ \hline
\multicolumn{1}{|l|}{\textbf{Identifier}} & \multicolumn{3}{l|}{\textbf{Quasi-Identifiers}}                                                                     & \multicolumn{2}{l|}{\textbf{Private attributes}}                                      & \textbf{Model prediction} \\ \hline
\multicolumn{1}{|l|}{Name}                & \multicolumn{1}{l|}{Age}         & \multicolumn{1}{l|}{Gender}     & \multicolumn{1}{l|}{City}                      & \multicolumn{1}{l|}{Salary}                & \multicolumn{1}{l|}{Relationship status} & Credit decision           \\ \hline
\multicolumn{1}{|l|}{\textit{Lisa}}      & \multicolumn{1}{l|}{\textit{\textbf{21}}} & \multicolumn{1}{l|}{\textit{F}} & \multicolumn{1}{l|}{\textit{\textbf{Brussels}}} & \multicolumn{1}{l|}{\textit{\textbf{\$50K}}} & \multicolumn{1}{l|}{\textit{Single}}     & \textit{Reject}           \\ \hline
\end{tabular}
\end{table}

\hspace{4cm} \arrowdown
\hspace{1cm} \small \parbox{8cm}{\hspace{1cm}\textbf{Counterfactual explanation}= \\If you would be \textbf{three years older}, lived in \textbf{Antwerp} and your income would be \textbf{\$10K} higher, you would have received a positive credit decision}

\begin{table}[h]
\centering
\begin{tabular}{|lllllll|}
\hline
\multicolumn{7}{|l|}{\textbf{Counterfactual instance}}                                                                                                                                                                                                                              \\ \hline
\multicolumn{1}{|l|}{\textbf{Identifier}} & \multicolumn{3}{l|}{\textbf{Quasi-Identifiers}}                                                                     & \multicolumn{2}{l|}{\textbf{Private attributes}}                                      & \textbf{Model prediction} \\ \hline
\multicolumn{1}{|l|}{Name}                & \multicolumn{1}{l|}{Age}         & \multicolumn{1}{l|}{Gender}     & \multicolumn{1}{l|}{City}                      & \multicolumn{1}{l|}{Salary}                & \multicolumn{1}{l|}{Relationship status} & Credit decision           \\ \hline
\multicolumn{1}{|l|}{\textit{Fiona}}      & \multicolumn{1}{l|}{\textit{\textbf{24}}} & \multicolumn{1}{l|}{\textit{F}} & \multicolumn{1}{l|}{\textit{\textbf{Antwerp}}} & \multicolumn{1}{l|}{\textit{\textbf{\$60K}}} & \multicolumn{1}{l|}{\textit{Single}}     & \textit{Accept}           \\ \hline
\end{tabular}\\
\captionof{figure}{Example of a counterfactual explanation}\label{cf_explanation} 
\end{table}

At the same time, there is a growing concern about the potential privacy risks of machine learning~\citep{liu2021machine}. Privacy is recognized as a human right and defined by Oxford Dictionary as \emph{“a state of being free from the attention of the public”}.\footnote{\url{https://www.oxfordlearnersdictionaries.com/definition/american_english/privacy}} 
In a privacy attack, the goal of an adversary is to gain knowledge that was not intended to be shared~\citep{liu2021machine,rigaki2020survey}. Different kinds of privacy attacks exist: 
both the training data, where the adversary tries to infer membership in a \emph{membership inference attack} or specific attributes of an input sample in an \emph{attribute inference attack}, as well as the model, in a \emph{model extraction attack}, can be the target~\citep{fredrikson2015model,rigaki2020survey}. 

Unfortunately, there exists an inherent tension between explainability and privacy as the usage of Explainable AI can increase these privacy risks~\citep{aivodji2020model}: model explanations offer users information about how the model made a decision about their data instance. Consequently, they leak information about the model and the data instances that were used to train the model. 
In this paper, we introduce a new kind of privacy attack based on counterfactual explanations and we call this an \emph{explanation linkage attack}. A \emph{linkage attack} attempts to identify anonymized individuals by combining the data with background information. An \emph{explanation} linkage attack attempts to link the counterfactual explanation with background information to identify the counterfactual instance. We illustrate an example of an explanation linkage attack in Section~\ref{sec:problem}.
Unfortunately, the introduction of these attacks indicates that an attempt to make an AI system safer by making it more transparent can have the opposite effect~\citep{sokol2019counterfactual}. Other researchers~\citep{budig2021trade,patel2020model} also confirm the trade-off between privacy and explainability and emphasise that assessing this trade-off for minority groups is an important direction for future research~\citep{patel2020model}.

Our contributions are as follows:
\begin{itemize}
    \item We introduce a new kind of privacy attack, the \emph{explanation linkage attack}, that can occur when using counterfactual explanations that are grounded in instances from the training set.
    \item As a solution for this problem, we propose $k$-anonymous counterfactual explanations and develop an algorithm to generate these.
    \item We evaluate how $k$-anonymizing the counterfactual explanations influences the quality of these explanations, and introduce \emph{pureness} as a new metric to evaluate the validity of these explanations.
    \item We show the trade-off between \emph{transparency}, \emph{fairness} and \emph{privacy} when using $k$-anonymous explanations: when we add more privacy constraints, the quality of the explanations and thus the transparency decreases. This effect on the explanation quality is larger for minority groups, as they tend to be harder to anonymize, and this can impact the fairness.
\end{itemize}

\section{Problem statement} \label{sec:problem}
We introduce the privacy problem of counterfactual explanations that are grounded in instances of the training set, and illustrate this problem by using a simple toy dataset.
 This dataset contains individuals that are described by a set of identifiers, quasi-identifiers and private attributes~\citep{sweeney2002k}. Identifiers are attributes such as name, phone or social security number and need to be suppressed in any case as they often do not have predictive value and can uniquely identify a person. Quasi-identifiers are attributes such as age, zip code or gender that can hold some predictive value. They are assumed to be public information; however, even though they cannot uniquely identify a person, their combination might. It has been shown that 87\% of US citizens can be re-identified by the combination of their zip code, gender and date of birth~\citep{sweeney2000simple}.  Private attributes are attributes that are not publicly known. 
We assume that an adversary will try to get access to the private attributes of a user in the dataset, and a possible avenue to achieve this goal, is by asking for counterfactual explanations. 
Assume the following factual instance \emph{Lisa} 
 in Table~\ref{factual_instance}:
 
\begin{table}[h]
\centering
\begin{tabular}{lllllll}
\hline
\multicolumn{1}{|l|}{\textbf{Identifier}}  &
\multicolumn{3}{|l|}{\textbf{Quasi-identifiers}}                                         & \multicolumn{2}{|l|}{\textbf{Private attributes}}                                          & \multicolumn{1}{|l|}{\textbf{Model prediction}}           \\ \hline
\multicolumn{1}{|l|}{Name}         &
\multicolumn{1}{|l|}{Age}         & \multicolumn{1}{l|}{Gender}     & \multicolumn{1}{l|}{City}              & \multicolumn{1}{l|}{Salary}       & \multicolumn{1}{l|}{Relationship status} & \multicolumn{1}{l|}{Credit decision} \\ \hline
\multicolumn{1}{|l|}{\textit{Lisa}} &
\multicolumn{1}{|l|}{\textit{21}} & \multicolumn{1}{l|}{\textit{F}} & \multicolumn{1}{l|}{\textit{Brussels}} & \multicolumn{1}{l|}{\textit{\$50K}} & \multicolumn{1}{l|}{\textit{Single}}     & \multicolumn{1}{l|}{\textit{Reject}} \\ \hline
\end{tabular}
\caption{Factual instance \emph{Lisa}}
\label{factual_instance}
\end{table}
$Name$ is the identifier that is deleted from the dataset, but, as mentioned, people can often be identified by their unique combination of quasi-identifiers. 
$Age$, $Gender$ and $City$ are the quasi-identifiers in this dataset that are assumed to be public knowledge for every adversary. A possible reasoning behind this, is that the adversary acquired access to a voter registration list as in \citet{sweeney2000simple}. $Salary$ and $Relationship$ are private attributes that one does not want to be public information, and the target attribute in this dataset is whether the individual will be awarded credit or not. \emph{Lisa} is predicted by the machine learning model as not creditworthy and her credit gets rejected. Logically, \emph{Lisa} wants to know the easiest way to get her credit application accepted, so she asks for a \emph{counterfactual explanation}, the smallest change to her feature values that result in a different prediction outcome.  

\begin{table}[h]
\centering
\begin{tabular}{lllllll}
\hline
\multicolumn{1}{|l|}{\textbf{Identifier}}  &
\multicolumn{3}{|l|}{\textbf{Quasi-identifiers}}                                         & \multicolumn{2}{|l|}{\textbf{Private attributes}}                                          & \multicolumn{1}{|l|}{\textbf{Model prediction}}           \\ \hline
\multicolumn{1}{|l|}{Name}         & 
\multicolumn{1}{|l|}{Age}         & \multicolumn{1}{l|}{Gender}     & \multicolumn{1}{l|}{City}              & \multicolumn{1}{l|}{Salary}        & \multicolumn{1}{l|}{Relationship status} & \multicolumn{1}{l|}{Credit decision} \\ \hline
\multicolumn{1}{|l|}{\textit{Alfred}} &
\multicolumn{1}{|l|}{\textit{25}} & \multicolumn{1}{l|}{\textit{M}} & \multicolumn{1}{l|}{\textit{Brussels}} & \multicolumn{1}{l|}{\textit{\$50K}}  & \multicolumn{1}{l|}{\textit{Single}}     & \multicolumn{1}{l|}{\textit{Reject}} \\ \hline
\multicolumn{1}{|l|}{\textit{Boris}} &
\multicolumn{1}{|l|}{\textit{23}} & \multicolumn{1}{l|}{\textit{M}} & \multicolumn{1}{l|}{\textit{Antwerp}}  & \multicolumn{1}{l|}{\textit{\$40K}}  & \multicolumn{1}{l|}{\textit{Separated}}  & \multicolumn{1}{l|}{\textit{Reject}} \\ \hline
\multicolumn{1}{|l|}{\textit{Casper}} &
\multicolumn{1}{|l|}{\textit{34}} & \multicolumn{1}{l|}{\textit{M}} & \multicolumn{1}{l|}{\textit{Brussels}} & \multicolumn{1}{l|}{\textit{\$30K}}  & \multicolumn{1}{l|}{\textit{Cohabiting}} & \multicolumn{1}{l|}{\textit{Reject}} \\ \hline
\multicolumn{1}{|l|}{\textit{Derek}} &
\multicolumn{1}{|l|}{\textit{47}} & \multicolumn{1}{l|}{\textit{M}} & \multicolumn{1}{l|}{\textit{Antwerp}}  & \multicolumn{1}{l|}{\textit{\$100K}} & \multicolumn{1}{l|}{\textit{Married}}    & \multicolumn{1}{l|}{\textit{Accept}} \\ \hline
\multicolumn{1}{|l|}{\textit{Edward}} &
\multicolumn{1}{|l|}{\textit{70}} & \multicolumn{1}{l|}{\textit{M}} & \multicolumn{1}{l|}{\textit{Brussels}} & \multicolumn{1}{l|}{\textit{\$90K}}  & \multicolumn{1}{l|}{\textit{Single}}     & \multicolumn{1}{l|}{\textit{Accept}} \\ \hline
\rowcolor{Gray}
\multicolumn{1}{|l|}{\textit{Fiona}} &
\multicolumn{1}{|l|}{\textit{24}} & \multicolumn{1}{l|}{\textit{F}} & \multicolumn{1}{l|}{\textit{Antwerp}}  & \multicolumn{1}{l|}{\textit{\$60K}}  & \multicolumn{1}{l|}{\textit{Single}}     & \multicolumn{1}{l|}{\textit{Accept}} \\ \hline
\multicolumn{1}{|l|}{\textit{Gina}} &
\multicolumn{1}{|l|}{\textit{27}} & \multicolumn{1}{l|}{\textit{F}} & \multicolumn{1}{l|}{\textit{Antwerp}}  & \multicolumn{1}{l|}{\textit{\$80K}}  & \multicolumn{1}{l|}{\textit{Married}}    & \multicolumn{1}{l|}{\textit{Accept}} \\ \hline
\multicolumn{1}{|l|}{\textit{Hilda}} &
\multicolumn{1}{|l|}{\textit{38}} & \multicolumn{1}{l|}{\textit{F}} & \multicolumn{1}{l|}{\textit{Brussels}} & \multicolumn{1}{l|}{\textit{\$60K}}  & \multicolumn{1}{l|}{\textit{Widowed}}    & \multicolumn{1}{l|}{\textit{Reject}} \\ \hline
\multicolumn{1}{|l|}{\textit{Ingrid}} &
\multicolumn{1}{|l|}{\textit{26}} & \multicolumn{1}{l|}{\textit{F}} & \multicolumn{1}{l|}{\textit{Antwerp}}  & \multicolumn{1}{l|}{\textit{\$60K}}  & \multicolumn{1}{l|}{\textit{Single}}   & \multicolumn{1}{l|}{\textit{Reject}} \\ \hline
\multicolumn{1}{|l|}{\textit{Jade}} &
\multicolumn{1}{|l|}{\textit{50}} & \multicolumn{1}{l|}{\textit{F}} & \multicolumn{1}{l|}{\textit{Brussels}} & \multicolumn{1}{l|}{\textit{\$100K}} & \multicolumn{1}{l|}{\textit{Married}}    & \multicolumn{1}{l|}{\textit{Accept}} \\ \hline
\end{tabular}
\caption{Training set}
\label{cf_table}
\end{table}
In our set-up, the counterfactual algorithm  looks for the instance in the training set that is nearest to \emph{Lisa} and has a different prediction outcome (the \emph{nearest unlike neighbor)}. 
The training set, with the nearest unlike neighbor highlighted, is shown in Table~\ref{cf_table}.
\emph{Fiona} has similar attribute values as \emph{Lisa}, but is 24 years old instead of 21, lives in Antwerp instead of Brussels and earns \$60K instead of \$50K.
When \emph{Fiona} is used as counterfactual instance by the explanation algorithm, \emph{Lisa} would receive the explanation: \emph{‘If you would be 3 years older, lived in Antwerp and your income was \$10K higher, then you would have received the loan'}. Based on her combined knowledge of the explanation and her own attributes, \emph{Lisa} can now deduce that $Fiona$ is the counterfactual instance, as there is only one person in this dataset with this combination of quasi-identifiers (a 24-year old woman living in Antwerp). Therefore, \emph{Lisa} can deduce the private attributes of \emph{Fiona}, namely \emph{Fiona}'s income and relationship status, which is undesirable. 

Obviously, this is just a toy example, but we envision many real-world settings where this situation could occur. For instance, when end users receive a negative decision, made by a \emph{high-risk AI system}: these systems are defined by the EU's AI Act, which categorizes the risk of AI systems usage into four levels~\citep{EUAiAct2022}. Among others, they include employment, educational training, law enforcement, migration and essential public services such as credit scoring. Article 13(1) states: \emph{``High-risk AI systems shall be designed and developed in such a way to ensure that their operation is sufficiently transparent to enable users to interpret the system's output and use it appropriately.''} 
These systems are thus obliged to provide some form of transparency and guidance to its users, which could be done by providing counterfactual explanations or any other transparency technique. Most of these settings use private attributes as input for their decisions, so it is important to make sure that the used transparency techniques do not reveal private information about other decision subjects. For example, in decisions about educational training or employment, someone's grades could be revealed, or in credit scoring,  the income of other decision subjects could be disclosed.

This privacy risk only occurs when the counterfactual algorithm uses instance-based strategies to find the counterfactual explanations. These counterfactuals correspond to the \emph{nearest unlike neighbor} and are also called \emph{native counterfactuals}~\citep{brughmans2021nice,keane2020good}. Other counterfactual algorithms use perturbation where synthetic counterfactuals are generated by perturbing the factual instance and labelling it with the machine learning model, without reference to known cases in the training set~\citep{keane2020good}.
These techniques are also vulnerable to privacy attacks such as model extraction but we focus on counterfactual algorithms that return real instances: several algorithms do this, as this substantially decreases the run time while also increasing desirable properties of the explanations such as plausibility~\citep{brughmans2021nice}. Plausibility measures how realistic the counterfactual explanation is with respect to the data manifold, which is a desirable property\citep{guidotti2022counterfactual}, and \citet{brughmans2021nice} show that the techniques resulting in an actual instance have the best plausibility results. Furthermore, it is argued that counterfactual instances that are plausible, are more robust and thus are less vulnerable to the uncertainty of the classification model or changes over time~\citep{artelt2021evaluating,brughmans2021nice,pawelczyk2020counterfactual}.  This shows that for some use cases it can be very useful to use real data points as counterfactuals instead of synthetic ones as for the latter the risk of generating implausible counterfactual explanations can be quite high~\citep{laugel2019dangers}.
 Algorithms that use these \emph{native} counterfactual explanations include NICE (without optimization setting)~\citep{brughmans2021nice}, the WIT tool with NNCE~\citep{wexler2019if}, FACE~\citep{poyiadzi2020face} and certain settings of CBR~\citep{keane2020good}.

\section{Proposed solution}
As a solution, we propose to make the counterfactual explanations \emph{$k$-anonymous}.
\emph{$k$-anonymity} is a property that captures the protection of released data against possible reidentification by stating that the released data should be indistinguishable between $k$ data subjects~\citep{van2014encyclopedia}.

\subsection{What is $k$-anonymity?} \label{sec:k-anom}
Before $k$-anonymity was introduced, data that looked anonymous was often freely shared after removing explicit identifiers such as name and address, incorrectly believing that individuals in those datasets could not be identified. Contrary to these beliefs, we have seen that people can often be identified through their unique combination of quasi-identifiers. 

Consider a database that holds private information about individuals, where each individual is described by a set of identifiers, quasi-identifiers, and private attributes. $k$-anonymity characterises the degree of privacy, where the information for each person in the dataset cannot be distinguished from at least $k-1$ other individuals whose information was also released~\citep{sweeney2002achieving}. A group of individuals that cannot be distinguished from each other and thus have the same values of quasi-identifiers are named an \emph{equivalence class}.


Usually $k$-anonymity is applied on the whole dataset: the quasi-identifiers of the data records are suppressed or generalised in such a way that one record is not distinguishable from at least $k-1$ other data records in that dataset~\citep{meyerson2004complexity}. In this way, the privacy of individuals is protected to some extent by \emph{“hiding in the crowd"} as private data can now only be linked to a set of individuals of at least size $k$~\citep{gionis2008k}. However, by generalising or suppressing attribute values, the data becomes less useful, so the problem studied is to make a dataset $k$-anonymous with minimal loss of information~\citep{gionis2008k, xu2006utility}. We will measure the loss in information value with the \emph{Normalized Certainty Penalty} (NCP) and explain this metric in Section~\ref{sec:eval_metrics}.

\subsection{Application to our problem}

\begin{table}[ht]
\centering
\begin{tabular}{|l|ll|}
\hline
                            & \multicolumn{2}{c|}{\textbf{$k$-anonymity}}                                                                                                                                                                                                                                                                                                                                      \\ \hline
                            & \multicolumn{1}{l|}{\textbf{Dataset}}                                                                                                                                                             & \textbf{Counterfactual explanation}                                                                                                                                        \\ \hline
\textbf{Input}              & \multicolumn{1}{l|}{Dataset}                                                                                                                                                                      & \begin{tabular}[c]{@{}l@{}}Dataset\\ Factual instance\\ Counterfactual explanation\\ Machine learning model\end{tabular}                                                                         \\ \hline
\textbf{Defined over}       & \multicolumn{1}{l|}{Dataset}                                                                                                                                                                      & Counterfactual explanation                                                                                                                                                 \\ \hline
\textbf{Method}             & \multicolumn{1}{l|}{Mondrian\footnotemark, Datafly\footnotemark,..}                                                                                                                                                         & CF-K                                                                                                                                                                      \\ \hline
\textbf{Risk}               & \multicolumn{1}{l|}{\begin{tabular}[c]{@{}l@{}}Identifying instances in the dataset \\ based on their combination \\ of quasi-identifiers and inferring\\  their private attributes\end{tabular}} & \begin{tabular}[c]{@{}l@{}}Identifying the counterfactual instance\\ based on its combination\\  of quasi-identifiers and inferring\\  its private attributes\end{tabular} \\ \hline
\textbf{Evaluation metrics} & \multicolumn{1}{l|}{\begin{tabular}[c]{@{}l@{}}Degree of privacy\\ Information loss\end{tabular}}                                                                                                 & \begin{tabular}[c]{@{}l@{}}Degree of privacy\\ Information loss\\ Counterfactual validity\end{tabular}                                                                     \\ \hline
\end{tabular}
\caption{Comparison between the original problem setting of $k$-anonymity and our problem setting.}
\label{tab:comparison}
\end{table}
\footnotetext[1]{\citet{lefevre2006mondrian}}
\footnotetext[2]{\citet{sweeney2002achieving}}
Our application differs from the original set-up of $k$-anonymity as it is focused on making counterfactual explanations anonymous and not the whole dataset. The original application has to be used in situations where the whole dataset is made public. We highlight this difference in Table~\ref{tab:comparison}.
A counterfactual instance is defined as $k$-anonymous if the combination of quasi-identifiers can belong to at least $k$ individuals in the training set, and consequently, a counterfactual explanation is defined as $k$-anonymous if the counterfactual instance on which it is based, is $k$-anonymous. 
We implement this by looking for close neighbours of \emph{Fiona}, that have similar values of quasi-identifiers, and that also have the desired prediction outcome. In this case, the closest neighbor to \emph{Fiona} that has the desired prediction outcome is \emph{Gina}, as can be seen in Table~\ref{cf_table}. Next, we generalise the quasi-identifiers of the counterfactual instance so that they can belong to both the counterfactual instance and the neighbour, resulting in a counterfactual instance that is at least 2-anonymous (see Figure~\ref{gen_instance2}.) However, by doing so we degrade the quality of the data as we will see in Section~\ref{sec:eval_metrics}.




\begin{table}[h]
\centering
\begin{tabular}{|lllllll|}
\hline
\multicolumn{7}{|l|}{\textbf{Counterfactual instance}}                                                                                                                                                                                                                              \\ \hline
\multicolumn{1}{|l|}{\textbf{Identifier}} & \multicolumn{3}{l|}{\textbf{Quasi-Identifiers}}                                                                     & \multicolumn{2}{l|}{\textbf{Private attributes}}   &    
\multicolumn{1}{l|}{\textbf{Model prediction}} \\ \hline
\multicolumn{1}{|l|}{Name}                & \multicolumn{1}{l|}{Age}         & \multicolumn{1}{l|}{Gender}     & \multicolumn{1}{l|}{City}                      & \multicolumn{1}{l|}{Salary}                & \multicolumn{1}{l|}{Relationship status} &  \multicolumn{1}{l|}{Credit decision}        \\ \hline
\multicolumn{1}{|l|}{\textit{*}}      & \multicolumn{1}{l|}{\textit{24}} & \multicolumn{1}{l|}{\textit{F}} & \multicolumn{1}{l|}{\textit{Antwerp}} & \multicolumn{1}{l|}{\textit{\$60K}} & \multicolumn{1}{l|}{\textit{Single}}  & 
\multicolumn{1}{l|}{\textit{Accept}}\\ \hline
\end{tabular}\\
\vspace{5mm}
{\Huge\textbf{+}}
\vspace{4mm}\\
\begin{tabular}{|lllllll|}
\hline
\multicolumn{7}{|l|}{\textbf{Neighbor}}                                                                                                                                                                                                                              \\ \hline
\multicolumn{1}{|l|}{\textbf{Identifier}} & \multicolumn{3}{l|}{\textbf{Quasi-Identifiers}}                                                                     & \multicolumn{2}{l|}{\textbf{Private attributes}}        &    
\multicolumn{1}{l|}{\textbf{Model prediction}}                               \\ \hline
\multicolumn{1}{|l|}{Name}                & \multicolumn{1}{l|}{Age}         & \multicolumn{1}{l|}{Gender}     & \multicolumn{1}{l|}{City}                      & \multicolumn{1}{l|}{Salary}                & \multicolumn{1}{l|}{Relationship status}
&  \multicolumn{1}{l|}{Credit decision} \\ \hline
\multicolumn{1}{|l|}{\textit{*}}      & \multicolumn{1}{l|}{\textit{27}} & \multicolumn{1}{l|}{\textit{F}} & \multicolumn{1}{l|}{\textit{Antwerp}} & \multicolumn{1}{l|}{\textit{\$80K}} & \multicolumn{1}{l|}{\textit{Married}}    & 
\multicolumn{1}{l|}{\textit{Accept}}       \\ \hline
\end{tabular}\\
\vspace{5mm}
\arrowdown 
\vspace{5mm}\\
\begin{tabular}{|lllllll|}
\hline
\multicolumn{7}{|l|}{\textbf{$K$-anonymous counterfactual instance}}                                                                                                                                                                                                                              \\ \hline
\multicolumn{1}{|l|}{Identifier} &
 \multicolumn{3}{|l|}{\textbf{Quasi-Identifiers}}                                                                     & \multicolumn{2}{l|}{\textbf{Private attributes}}   &    
\multicolumn{1}{l|}{\textbf{Model prediction}} \\ \hline
 \multicolumn{1}{|l|}{Name} &
\multicolumn{1}{|l|}{Age}         & \multicolumn{1}{l|}{Gender}     & \multicolumn{1}{l|}{City}                      & \multicolumn{1}{l|}{Salary}                & \multicolumn{1}{l|}{Relationship status}
&  \multicolumn{1}{l|}{Credit decision} \\ \hline
\multicolumn{1}{|l|}{\textit{*}} &
\multicolumn{1}{|l|}{\textit{24-27}} & \multicolumn{1}{l|}{\textit{F}} & \multicolumn{1}{l|}{\textit{Antwerp}} & \multicolumn{1}{l|}{\textit{\$60K}} & \multicolumn{1}{l|}{\textit{Single}} & 
\multicolumn{1}{l|}{\textit{Accept}}          \\ \hline
\end{tabular}

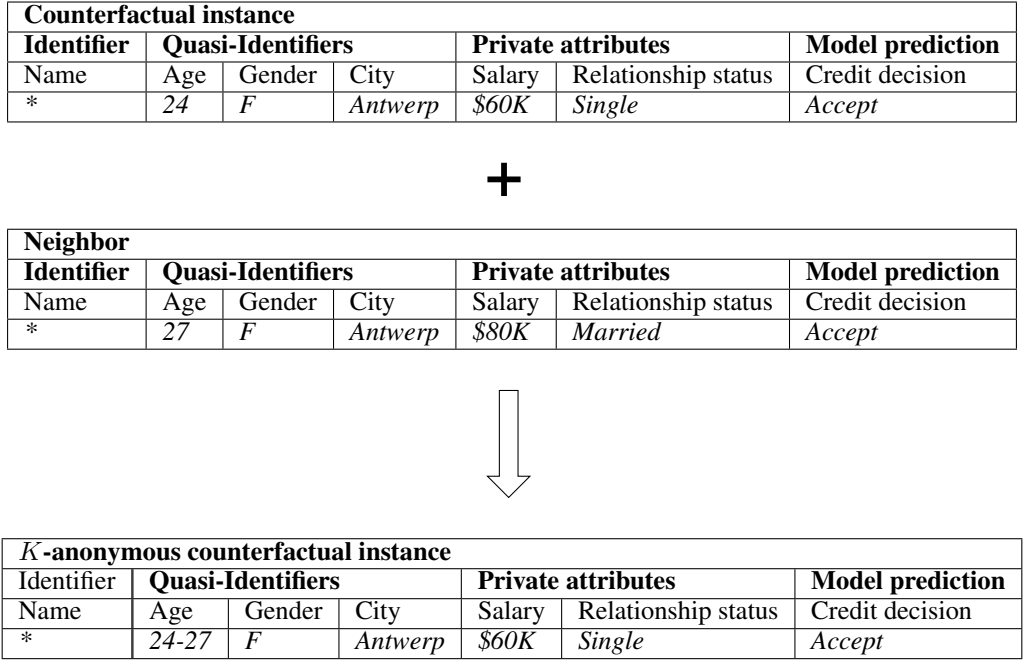
\captionof{figure}{How to generalize the counterfactual instance. As can be seen, we generalize only the values of the quasi-identifiers. The private attributes are still the same as in the original counterfactual instance as their attribute value is not public and therefore cannot be used to identify someone.}\label{gen_instance2} 
\end{table}
The $k$-anonymous counterfactual explanation based on the $k$-anonymous counterfactual instance in Figure~\ref{gen_instance2} and factual instance \emph{Lisa} (21, F, Brussels, \$50K, Single) is:
\emph{‘If you would be 3-6 years older, lived in Antwerp and had an income of \$60K, you would have received the loan'}.
This explanation is 3-anonymous because the combination of quasi-identifiers in the counterfactual instance (24-27, F, Antwerp) could point to at least three instances in the training set in Table~\ref{cf_table}, namely \emph{Fiona}, \emph{Gina} and \emph{Ingrid}.

However, the fact that other instances than the ones explicitly used to generate the $k$-anonymous counterfactual explanation, might also fall in the range of the explanation, introduces a new issue that is specific to $k$-anonymous counterfactual explanations. Counterfactual explanations are defined as the smallest change to the feature values of an instance that alter its prediction outcome, but does this still hold for $k$-anonymous counterfactual explanations? We are no longer sure that all the value combinations in the $k$-anonymous counterfactual instance lead to a change in the  prediction outcome and therefore we are not sure whether they are \emph{valid} counterfactual explanations.


In this toy example, the value combination of \emph{Ingrid} in Table~\ref{cf_table} is also part of the  $k$-anonymous counterfactual instance, as \emph{Ingrid} is between 24 and 27 years old, female, single, living in Antwerp and earning \$60K.
However, the model predicts \emph{Ingrids} credit decision to be rejected.
A possible reasoning behind this could be because the model has learned that for higher age groups a higher income is required to be awarded the credit (or any other pattern). Therefore,  if \emph{Lisa} would follow-up the \emph{“advice"} in the counterfactual explanation, it is possible that she would end up in this value combination, which does not result in an altering of the prediction outcome. This is problematic as this is one of the key objectives of counterfactual explanations.

This issue leads us to a new metric: how \emph{valid} is the $k$-anonymous counterfactual explanation?
We discuss the evaluation metrics further in Section~\ref{sec:eval_metrics}.

\section{Evaluation metrics} \label{sec:eval_metrics}
We measure the quality of the explanations by using the following metrics:
\begin{itemize}
    \item The degree of privacy is measured by $k$: to how many instances from the training set can this counterfactual explanation be linked?
    \item The validity of the counterfactual explanations is measured by the \emph{pureness}.
    \item The loss in information value is measured by  the \emph{Normalized Certainty Penalty (NCP)}.
\end{itemize}
We look at the interplay of these three metrics in Section~\ref{sec:trade-off}.
\paragraph{Degree of privacy} We measure the degree of privacy by using the definition of $k$-anonymity. In our toy example, $k$ is 3, as the generalised quasi-identifiers of the $k$-anonymous counterfactual instance could belong to three people when we look at the training set in Table~\ref{cf_table} (\emph{Fiona}, \emph{Gina} and \emph{Ingrid}).
 
\paragraph{Counterfactual validity}
We define a possible value combination as a combination of attribute values that is in the range of the $k$-anonymous counterfactual instance.
For a categorical attribute, we look at all the values present in the $k$-anonymous counterfactual instance. For a numerical attribute, we look at all the values that are in the range of the $k$-anonymous counterfactual instance \emph{and} are also present in the training set. We illustrate these calculations in Table~\ref{tab_val}. The pureness of a $k$-anonymous counterfactual explanation can be calculated as follows:

\begin{equation*}
    \text{Pureness} = \frac{\text{\# of value combinations with desired prediction outcome}}{\text{\# of value combinations}}
\end{equation*}

The theoretical pureness is calculated on all the value combinations, but we will approximate this by querying the model with 100 random combinations\footnote{We chose for 100 random value combinations instead of trying out all the possibilities as the number of combinations can quickly become very large when there is a lot of generalization. The more random value combinations we test, the more we approximate the theoretical pureness, but the longer the computation time.} and see how many of these combinations lead to the desired prediction outcome. The pureness is the proportion of these value combinations that lead to the desired prediction outcome, which obviously should be as high as possible (preferably 100\%).

\vspace{0.5cm}
Table~\ref{tab_val} shows all possible value combinations of the $k$-anonymous counterfactual instance, and the prediction outcome to each value combination. The goal of the counterfactual explanation was to alter the prediction outcome from \emph{Reject} to \emph{Accept}, so this is the desired prediction outcome.
The $k$-anonymous counterfactual explanation in our toy example leads to the desired prediction outcome in $50\%$ of the cases ($\frac{2}{4}$). If we sample 100 times out of the value combinations above, we expect this to approximate the theoretical pureness of 50\%.
\begin{table}[h]
\centering
\begin{tabular}{|l|l|l|l|l|l|}
\hline
\textbf{Age} & \textbf{Gender} & \textbf{City}    & \textbf{Salary} & \textbf{Relationship status} & \textbf{Model prediction} \\ \hline
\textit{24}  & \textit{F}      & \textit{Antwerp} & \textit{\$60K}    & \textit{Single}              & \textit{Accept}           \\ \hline
\textit{25}  & \textit{F}      & \textit{Antwerp} & \textit{\$60K}    & \textit{Single}              & \textit{Accept}           \\ \hline
\textit{26}  & \textit{F}      & \textit{Antwerp} & \textit{\$60K}    & \textit{Single}              & \textit{Reject}           \\ \hline
\textit{27}  & \textit{F}      & \textit{Antwerp} & \textit{\$60K}    & \textit{Single}              & \textit{Reject}           \\ \hline
\end{tabular}
\caption{Possible value combinations and its model predictions.} \label{tab_val}
\end{table}

\paragraph{Loss in information value}
 When datasets are made $k$-anonymous, they tend to lose information. In general, excessive anonymization makes the data less useful because some analysis is no longer possible or the analysis provides biased and incorrect results~\citep{el2008protecting}.  
A variety of metrics to measure this information loss have been proposed such as the Discernibility Metric (DM)~\citep{bayardo2005data}, Normalized Certainty Penalty (NCP)~\citep{xu2006utilityb} and Information Loss (IL)~\citep{lefevre2006workload,ghinita2009framework}. We use NCP as it has been approved as a suitable measure in the data anonymization literature~\citep{ghinita2007fast,tang2010utility,saeed2018anatomization}. NCP penalises attributes for the way they are generalised and captures the uncertainty caused by this generalization~\citep{xu2006utilityb}. It assigns larger penalties when attribute values are mapped to generalised values that replace many other values~\citep{loukides2012utility}. An advantage of this metric is that it can give different weights to different attributes, as some attributes can be more important than others for the data analysis process~\citep{xu2006utilityb}.
The NCP for each numerical (\emph{Num}) quasi-identifier $A$ in an equivalence class $G$ is defined as:
\begin{equation} \label{ncp_num}
    \text{NCP}_{A_{Num}} (G)=\frac{\max^G_{A_{Num}} - \min^G_{A_{Num}}}{\max_{A_{Num}}- \min_{A_{Num}}},
\end{equation}
where the numerator and denominator represent the range of attribute $A$ for the equivalence class $G$ and for the whole dataset respectively~\citep{ghinita2009framework}. This metric thus measures which part of the total range of the numerical attribute, is present in the equivalence class. Higher values signify more generalization, and consequently, more information loss.
In the case of a categorical (\emph{Cat}) quasi-identifier $A$, NCP is defined as follows:

\begin{equation} \label{ncp_cat}
    \text{NCP}_{A_{Cat}} (G) = \begin{cases}
        0, & \text{if $|A^G| = 1$}\\
        \frac{|A^G|}{|A|}, & \text{ otherwise}
    \end{cases} 
\end{equation}

where $|A|$ is the number of distinct values of attribute $A$ in the whole dataset, and $|A^G|$ is the number of distinct values of attribute $A$ in equivalence class $G$~\citep{xu2006utilityb}.
So, for a categorical attribute, this metric will check which proportion of possible unique values is present in the $k$-anonymous counterfactual instance. The higher this number is, the more generalized this attribute will be and the more information about this attribute is lost.
The NCP of equivalence class $G$ over all quasi-identifier attributes is:
\begin{equation} \label{ncp_tot}
    NCP(G) = \sum_{i=1}^d w_i \cdot NCP_{A_i} (G),
\end{equation}
where $d$ is the number of quasi-identifiers in the dataset, $A_i$ is a (numerical or categorical) attribute with weight $w_i$, where $\sum_i w_i =1 $~\citep{ghinita2009framework}. For our experiments, we assume all attributes have an equal weight but this can easily be altered in future experiments.
NCP measures the information loss for a single instance and its equivalence class. This can be aggregated to the information loss in the entire dataset~\citep{ghinita2009framework,xu2006utilityb} but for our problem setting, we only need to calculate the NCP for each $k$-anonymous counterfactual explanation, which constitutes one equivalence class.
As an illustration, we calculate the NCP of the $k$-anonymous counterfactual explanation (CE) in our toy example\footnote{See Table~\ref{cf_table} for the range of each attribute in the training set.}:

\begin{equation*} 
    NCP_{Age} (CE) =\frac{max_{{CE}^{Age}} - min_{{CE}^{Age}}}{max_{Age}- min_{Age}} = \frac{27-24}{70-23}= 0.064,
\end{equation*}
\begin{equation*} 
    \text{NCP}_{Gender} (CE) = 0 \qquad  (|A^{CE}| = 1),\\
    \text{NCP}_{City} (CE) = 0  \qquad  (|A^{CE}| = 1),
\end{equation*}
\begin{equation*} 
    \text{NCP}(CE) = \frac{1}{3} \cdot 0.064 + \frac{1}{3} \cdot 0 + \frac{1}{3} \cdot 0 = 0.021
\end{equation*}

\section{Materials and Methods}
\subsection{Materials}

\begin{table}[H]
\centering
\begin{tabular}{|l|l|l|l|l|l|}
\hline
\textbf{Dataset}  & \textbf{\# instances} & \textbf{\# attributes} & \textbf{QID}                                                                                                 & \textbf{\begin{tabular}[c]{@{}l@{}}Sensitive  \\ attribute\end{tabular}} & \textbf{\begin{tabular}[c]{@{}l@{}}Target  \\ attribute\end{tabular}}\\ \hline
\textbf{Adult\footnotemark}    & 48,842                 & 11                     & \textit{\begin{tabular}[c]{@{}l@{}}Age, Sex, Race,\\ Relationship,\\ Marital status\end{tabular}}            & \textit{Sex}   & \textit{Income}                                                            \\ \hline
\textbf{CMC\footnotemark}      & 1,473                  & 8                      & \textit{\begin{tabular}[c]{@{}l@{}}WifeAge,\\ ChildrenBorn\end{tabular}}                                     & \textit{WifeReligion}  & \textit{\begin{tabular}[c]{@{}l@{}}Contraceptive  \\ method\end{tabular}}                                                   \\ \hline
\textbf{German\footnotemark}   & 1,000                  & 19                     & \textit{\begin{tabular}[c]{@{}l@{}}Age, Foreign,\\ Personal status,\\ Residence time,\\ Employment, Job, \\ Property, Housing\end{tabular}}           & \textit{Personal status}                           & \textit{\begin{tabular}[c]{@{}l@{}}Credit  \\ decision\end{tabular}}                        \\ \hline
\textbf{Heart \footnotemark} & 303                   & 12                     & \textit{Age, Sex}                                                                                            & \textit{Sex}      & \textit{\begin{tabular}[c]{@{}l@{}}Heart  \\ disease\end{tabular}}                                                        \\ \hline
\textbf{Hospital\footnotemark} & 8,160       & 20            & \textit{\begin{tabular}[c]{@{}l@{}}Age Group, Race,\\ Gender, Ethnicity,\\ Zip Code - 3 digits\end{tabular}} & \textit{Gender}      & \textit{Costs}                                                      \\ \hline
\textbf{Informs\footnotemark}  & 5,000                  & 13                     & \textit{\begin{tabular}[c]{@{}l@{}}Dobmm, Dobyy,\\ Sex, Marry,\\ Educyear\end{tabular}}                      & \textit{Race}       & \textit{Income}                                                        \\ \hline
\end{tabular}
\caption{Description of used datasets with dataset properties} \label{used_datasets}
\end{table}
\footnotetext[6]{\url{https://github.com/EpistasisLab/pmlb/tree/master/datasets/adult}}
\footnotetext[7]{\url{https://archive.ics.uci.edu/ml/machine-learning-databases/cmc/}}
\footnotetext[8]{\url{https://github.com/EpistasisLab/pmlb/tree/master/datasets/german}}
\footnotetext[9]{\url{https://github.com/EpistasisLab/pmlb/tree/master/datasets/heart_c}}
\footnotetext[10]{\url{https://www.opendatanetwork.com/dataset/health.data.ny.gov/82xm-y6g8}}
\footnotetext[11]{\url{https://github.com/kaylode/$k$-anonymity/tree/main/data/informs}}

We choose the datasets described in Table~\ref{used_datasets}, as they are all tabular datasets that contain various personal attributes through which individuals could be identified, and are often used in research about privacy-preserving data mining~\citep{kisilevich2008kactus,simi2017extensive,slijepvcevic2021k}.\footnote{\url{https://github.com/kaylode/$k$-anonymity}} All these datasets contain private information such as financial and health data that people generally do not want to be made public.

\subsection{Methods}

\begin{figure}[H]
    \includegraphics[width=\textwidth]{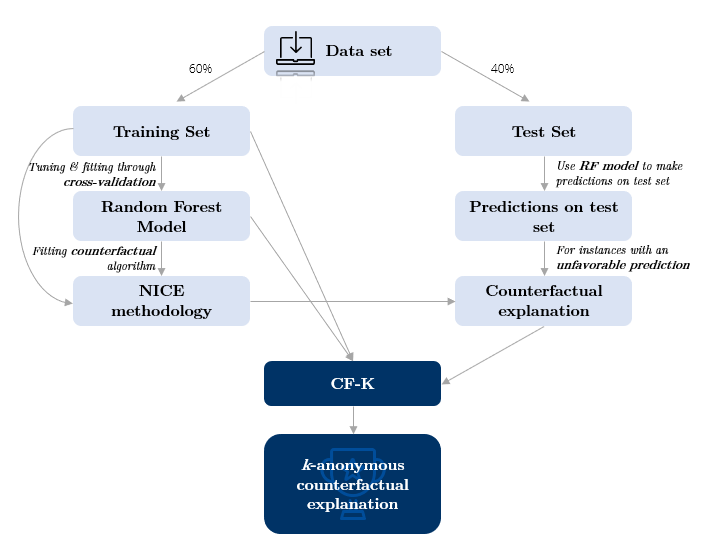}
    \caption{Used methodology to generate $k$ anonymous counterfactual explanations from a dataset.}
    \label{methodology}
\end{figure}

On every dataset, we apply the methodology as described in Figure~\ref{methodology}. We first split the dataset in a training and test set, using a split of 60-40. We fit and tune a Random Forest model through cross-validation on the training set. The following grid is used for tuning:
\begin{flalign*}
  \text{n\_estimators} &= [10, 50, 100, 500, 1,000, 5,000]  \\
\text{max\_leaf\_nodes} &= [10,100,500, n]    \text{ with $n=\infty$}
\end{flalign*}
We use the standard version (no optimization setting) of NICE~\citep{brughmans2021nice} as counterfactual algorithm and fit this on the training set and  the trained machine learning model. This trained machine learning model is used to make predictions on all the instances in the test set. For all the test instances\footnote{We set a limit at 1,000 instances for the sake of time.} without the desired prediction outcome, we use NICE to generate a counterfactual explanation. We focus on the test instances without the desired prediction outcome as these are the instances that generally use counterfactual explanations to receive advice on how to change their prediction outcome. As mentioned, when using NICE without any optimization setting, the counterfactual instances are real instances from the training data so they should be anonymized. The final step is to use CF-K to make these explanations $k$-anonymous.

\subsection{Novel algorithm CF-K}
In the original application of $k$-anonymity, the whole dataset is made public and thus has to be made $k$-anonymous. The goal is to find an optimal partition, for which both exact algorithms~\citep{lefevre2005incognito} and heuristics like genetic algorithms~\citep{iyengar2002transforming} and greedy algorithms~(Mondrian~\citep{lefevre2006mondrian}, Datafly~\citep{sweeney2002achieving}) exist. 

Our approach differs from the approaches published in the literature, as only the counterfactual explanation is made public and not the whole dataset. Therefore, we search for an equivalence class for each returned counterfactual instance separately. This changes the set-up of the problem, as making the whole dataset $k$-anonymous can degrade the data more than just making the counterfactual explanations $k$-anonymous: not every training instance is used as counterfactual explanation and unused training instances do not need to be made $k$-anonymous or used in the calculation for the best clustering. In the same way that local encoding achieves less information loss than global recoding~\citep{xu2006utilityb}, we hypothesize that only $k$-anonymizing the counterfactual instances can achieve lower information loss. We verify this claim in Section~\ref{sec:comparison}.
 Furthermore, specifically for our problem of $k$-anonymous counterfactual explanations we have to take the \emph{counterfactual validity} of the $k$-anonymous explanations into account as this is essential for the goal of counterfactual explanations. We use this as additional metric in our algorithm. 
 
 We name our algorithm that makes counterfactual explanations $k$-anonymous CF-K. It is based on the metaheuristic GRASP (Greedy Randomized Adaptive Search Procedure)~\citep{feo1995greedy}. GRASP is a multi-start metaheuristic in which each iteration consists of two phases: construction and local search. After the two phases, the current best solution is updated. The construction phase builds a feasible solution, and the local search phase searches the neighborhood until a local optimum is found~\citep{feo1995greedy}. In this construction phase, GRASP combines greediness with randomness, with the purpose of escaping the myopic behavior of a purely greedy algorithm.
 We choose a heuristic algorithm, as it is a NP-hard problem and we are not looking for the optimal solution but for the best solution that can be found in limited computing time.  Our aim is to provide a method that performs well, but we expect further optimizations to be possible in future research. More information about our implementation of these two phases and the choice of parameters can be found in the Appendix.

\section{Results}
\subsection{Results per dataset}

\begin{table}[h]
\centering
\begin{tabular}{|l|l|l|l|l|l|l|}
\hline
\textbf{Dataset}         & Adult            & CMC              & German           & Heart  & Hospital         & Informs      \\ \hline
\textbf{NCP  (mean)}     & 0.74\%  & 3.85\%  & 20.87\% & 2.64\% & 0.40\%  & 9.25\%  \\ \hline
\textbf{Pureness (mean)} & 99.79\% & 92.71\% & 98.41\% & 100\%  & 98.64\% & 92.26\% \\ \hline
\textbf{Execution time (mean)} & 13.1s & 10.8s & 14.8s & 2.1s  & 15.5s & 37.7s \\ \hline
\end{tabular}
\caption{Results of the average information loss (NCP) and counterfactual validity (pureness) over all the datasets. Algorithmic settings: $k = 10$, $\alpha = 20$, number of iterations~=~3.}
\end{table}

We see that the explanations of some datasets such as \emph{Adult} or \emph{Hospital} seem to be \emph{easier} to make private as their average information loss is a lot lower than in the other datasets. These are the two largest datasets, so arguably, it is easier to \emph{“hide in the crowd"}. \emph{Informs} is also a large dataset, but we see here that the combinations of quasi-identifiers more often uniquely identify a person so the explanations need to be generalized more to make them anonymous.
The \emph{German} dataset has the highest average NCP, which could be explained by it being a small dataset with the largest amount of quasi-identifiers. This is a pattern that has been observed in literature as well: when there are more attributes that can uniquely identify a person, the attribute values have to be generalized more~\citep{ayala2014systematic}.  We also see that in the \emph{Heart} dataset, the counterfactual validity measured by the pureness of a sample of 100 value combinations is always 100\%. We expect this to be the case if the quasi-identifiers, which are \emph{Age} and \emph{Sex} in this case, have a small influence on the outcome of the machine learning model. We verify this by examining the feature importance ranking of the used model, and indeed see that the quasi-identifiers are ranked very low. This could explain why generalizing them has no effect on the counterfactual validity. On the other hand, the \emph{CMC} dataset has the lowest average counterfactual validity, and here we see that the quasi-identifiers (\emph{WifeAge} and \emph{ChildrenBorn}) have a high value in the feature importance ranking. 

With regard to the exeuction time, we see that it will be fast enough for most applications, and is in line with the order of magnitude of generating counterfactual explanations~\citep{de2021framework}. If further speed-ups are necessary, this can be realised by decreasing the number of iterations, further optimization of the algorithm or using a stronger computer. All measurements were taken on a Dell Latitude 7400 laptop with 16GB of RAM and Intel\textregistered Core$^{TM}$ i7-8665U CPU. 

\subsection{Comparison with Mondrian} \label{sec:comparison}

\begin{figure}[h] 
\begin{subfigure}{0.48\textwidth}
\includegraphics[width=\linewidth]{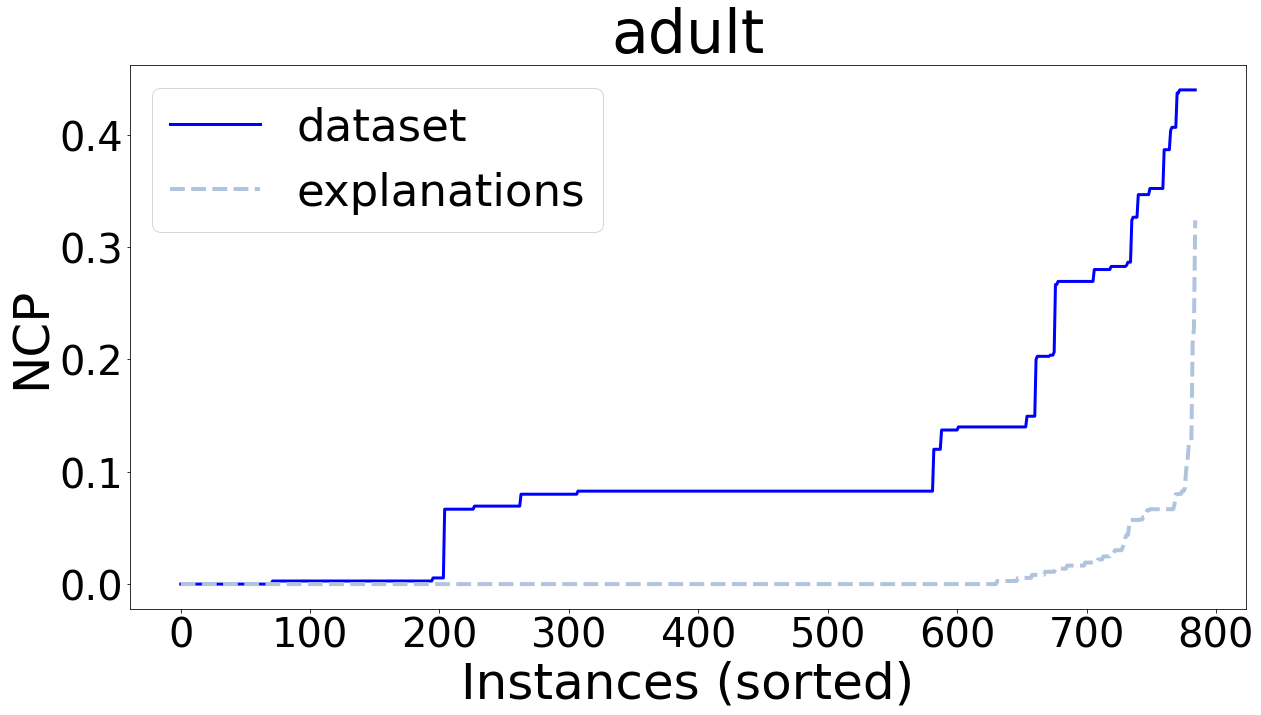}
\end{subfigure}\hspace*{\fill}
\begin{subfigure}{0.48\textwidth}
\includegraphics[width=\linewidth]{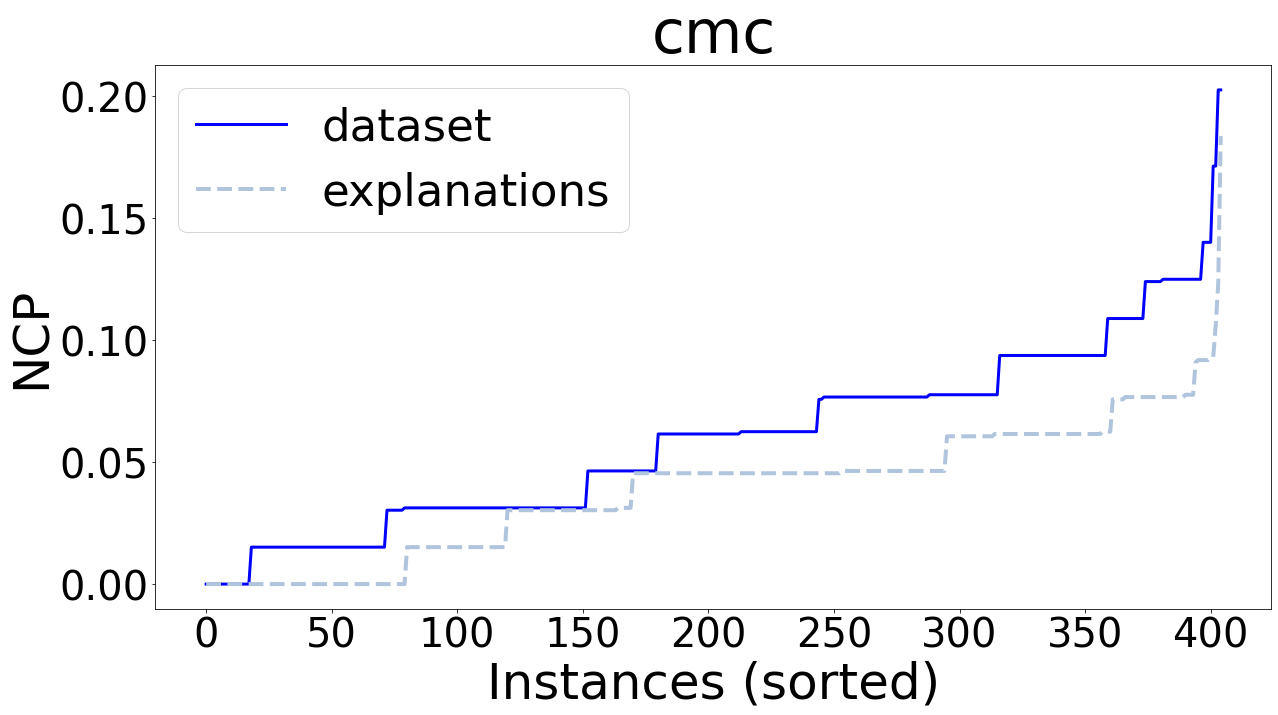}
\end{subfigure}
\medskip
\begin{subfigure}{0.48\textwidth}
\includegraphics[width=\linewidth]{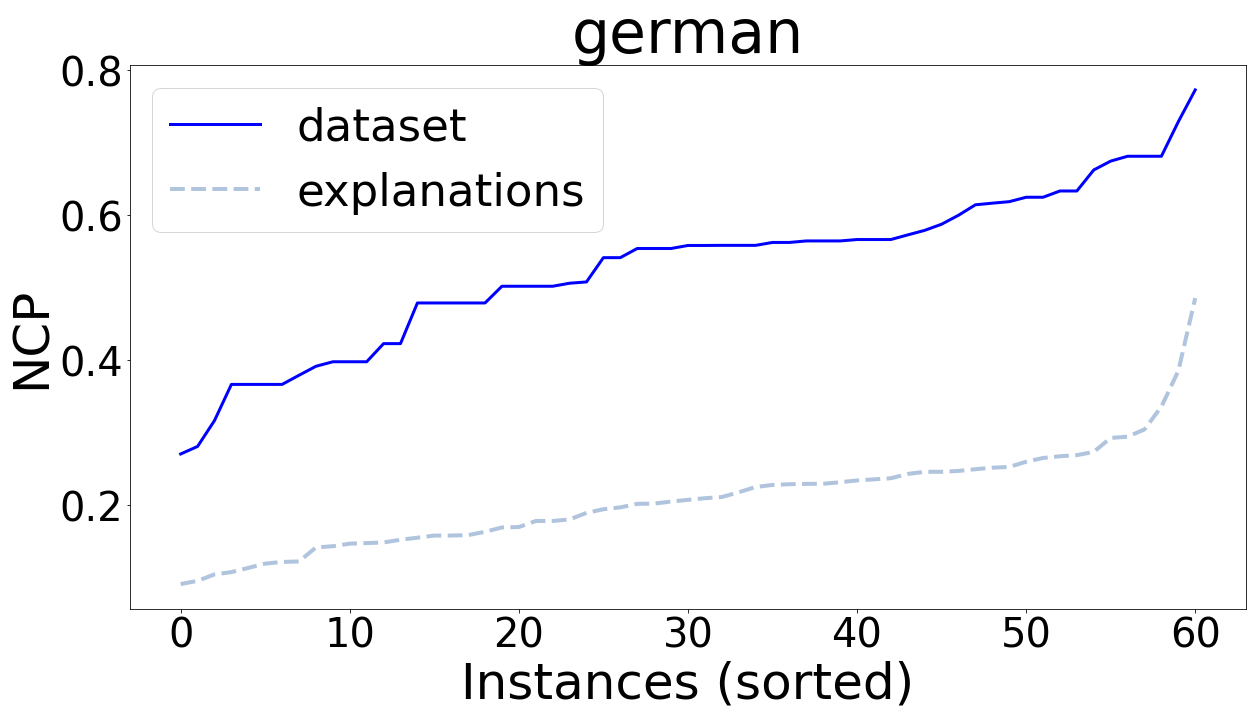}
\end{subfigure}\hspace*{\fill}
\begin{subfigure}{0.48\textwidth}
\includegraphics[width=\linewidth]{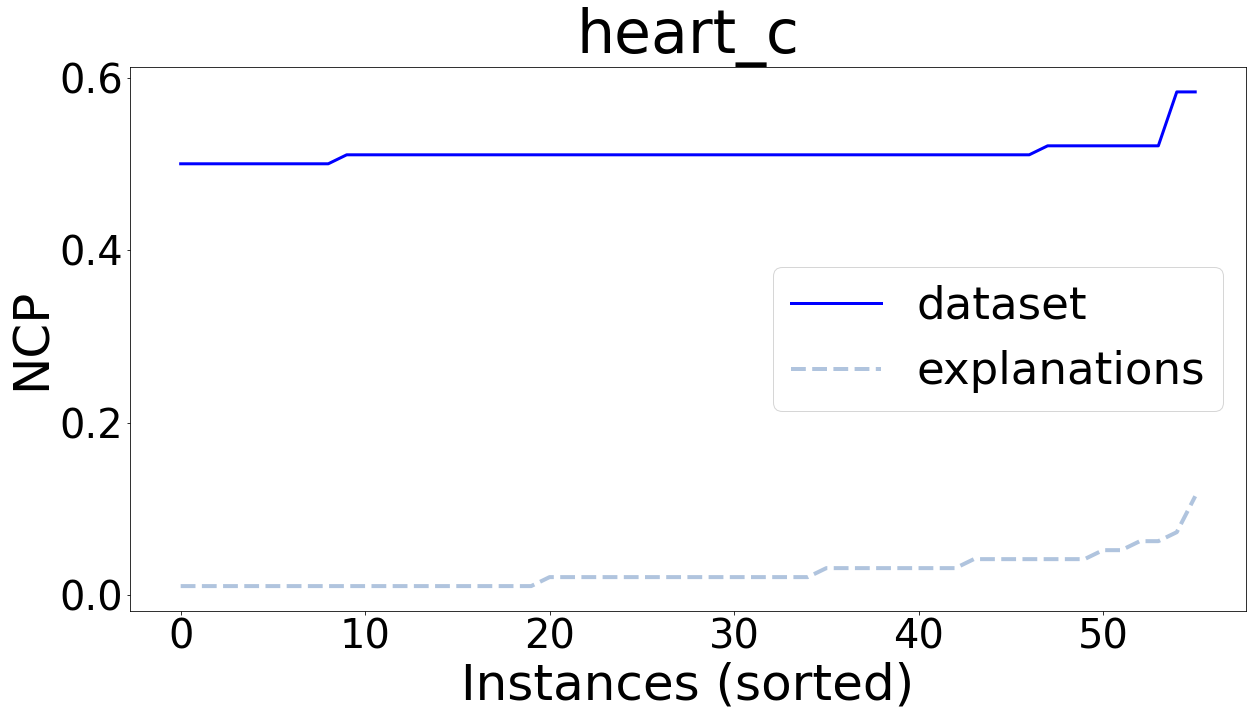}
\end{subfigure}
\medskip
\begin{subfigure}{0.48\textwidth}
\includegraphics[width=\linewidth]{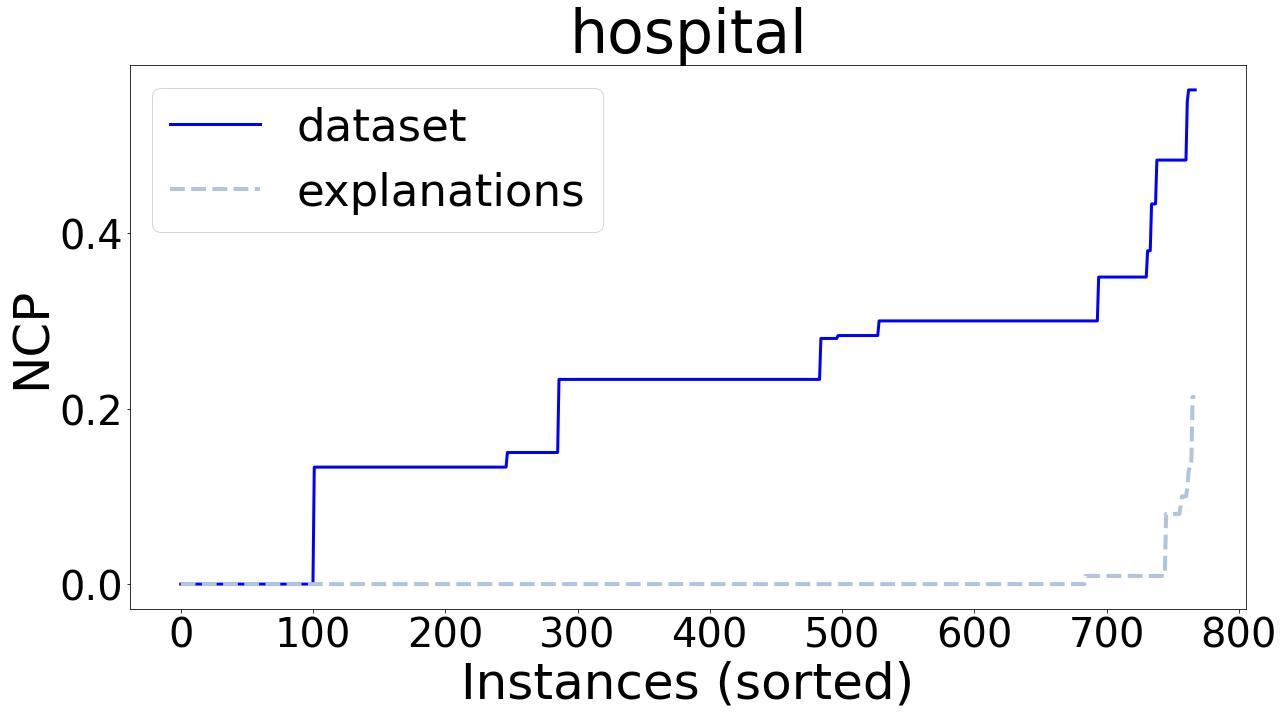}
\end{subfigure}
\begin{subfigure}{0.48\textwidth}
\includegraphics[width=\linewidth]{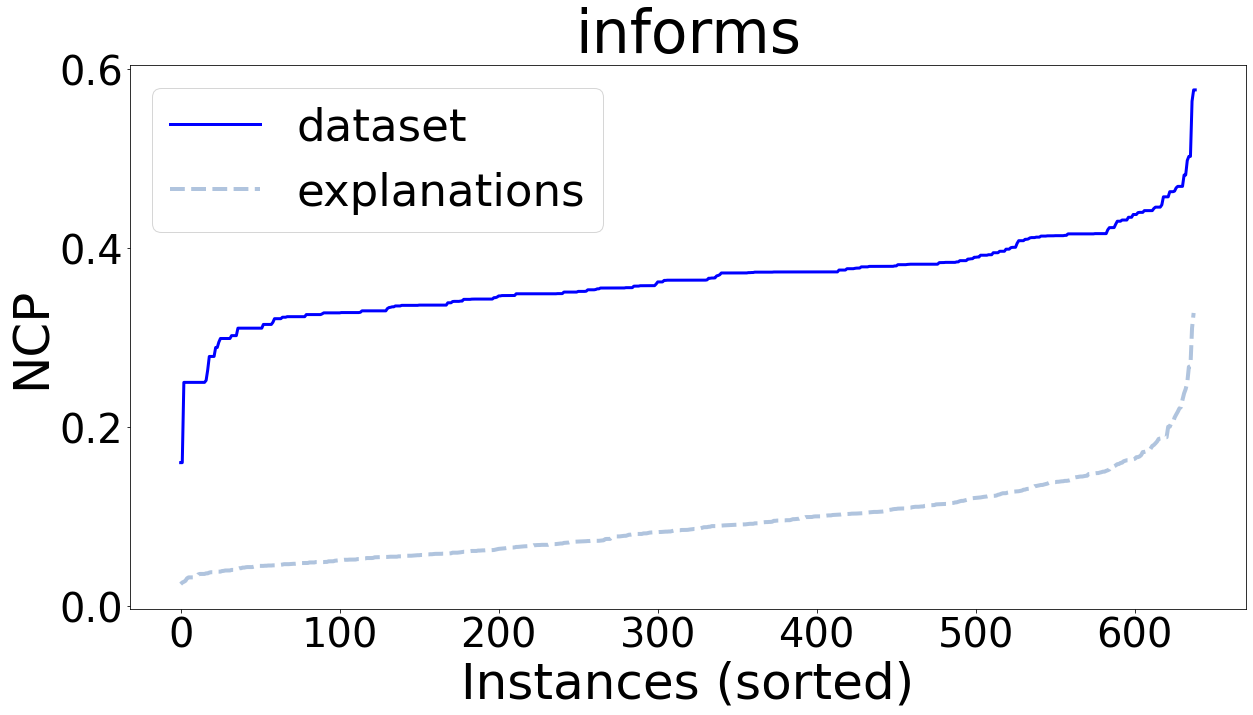}
\end{subfigure}\hspace*{\fill}
\caption{Comparison of NCP between Mondrian and CF-K with $k=10$. The instances are sorted by NCP.} \label{ncp_mondrian}
\end{figure}

\begin{figure}[ht] 
\begin{subfigure}{0.48\textwidth}
\includegraphics[width=\linewidth]{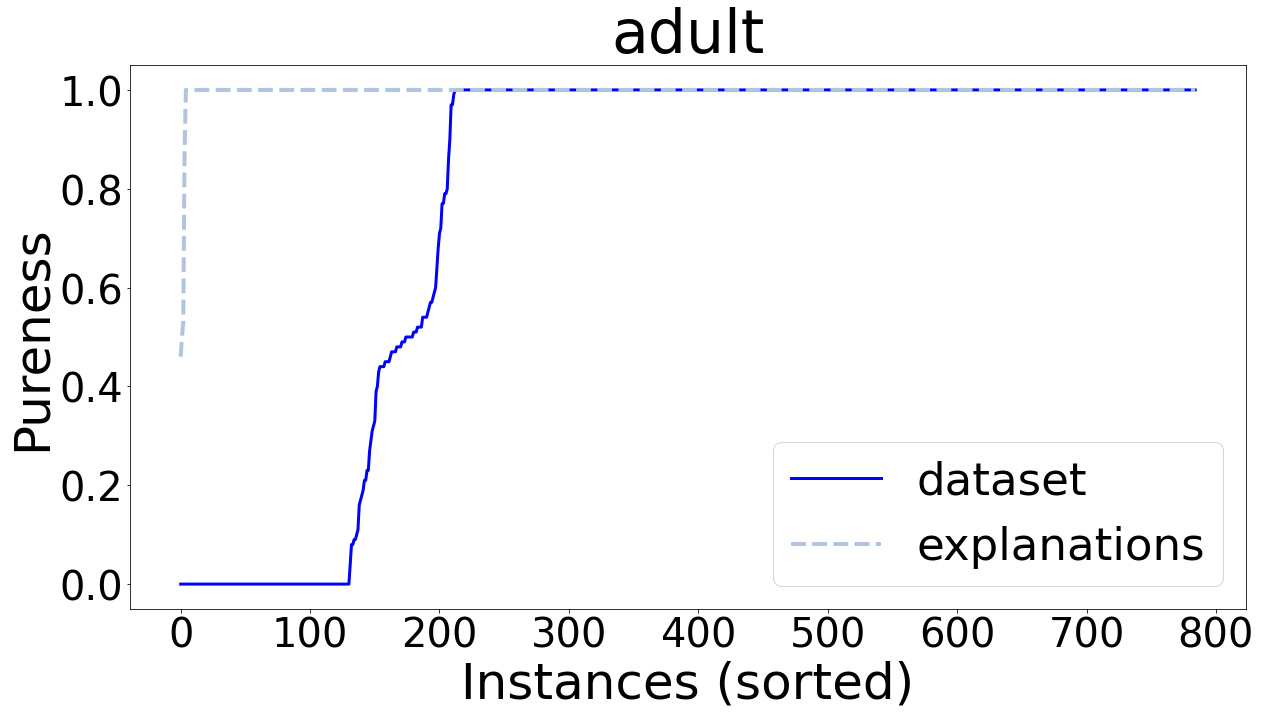}
\end{subfigure}\hspace*{\fill}
\begin{subfigure}{0.48\textwidth}
\includegraphics[width=\linewidth]{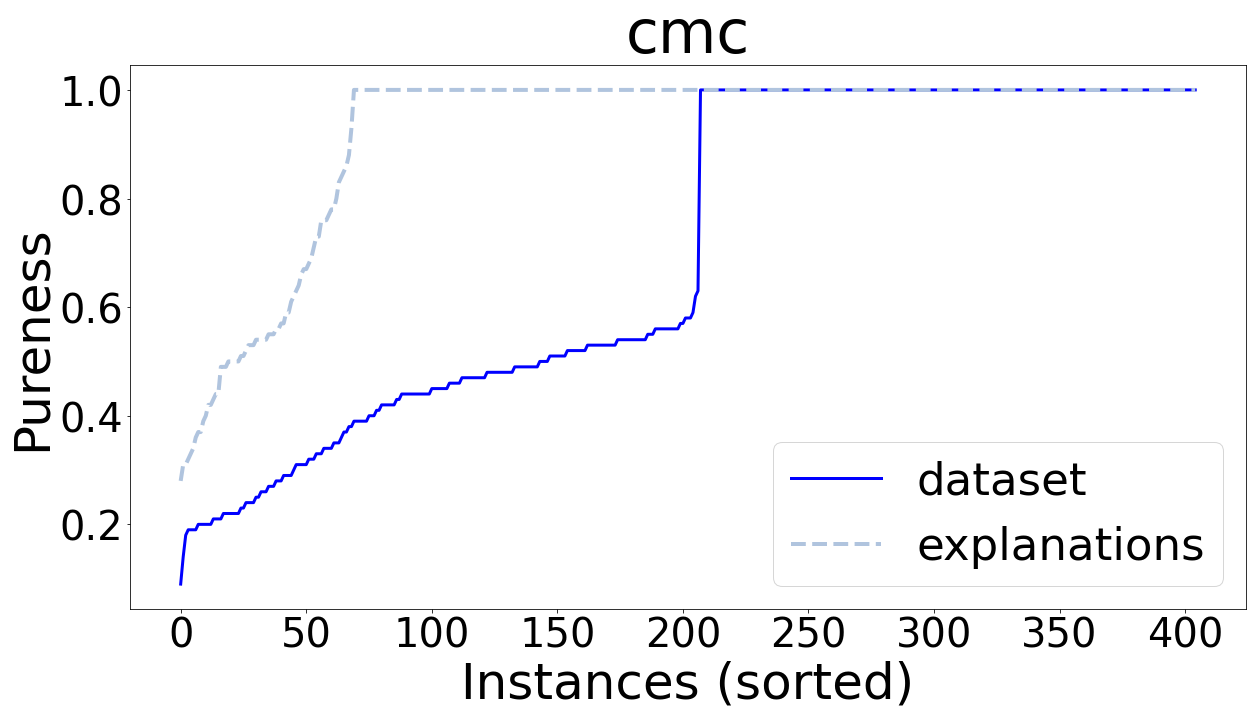}
\end{subfigure}
\medskip
\begin{subfigure}{0.48\textwidth}
\includegraphics[width=\linewidth]{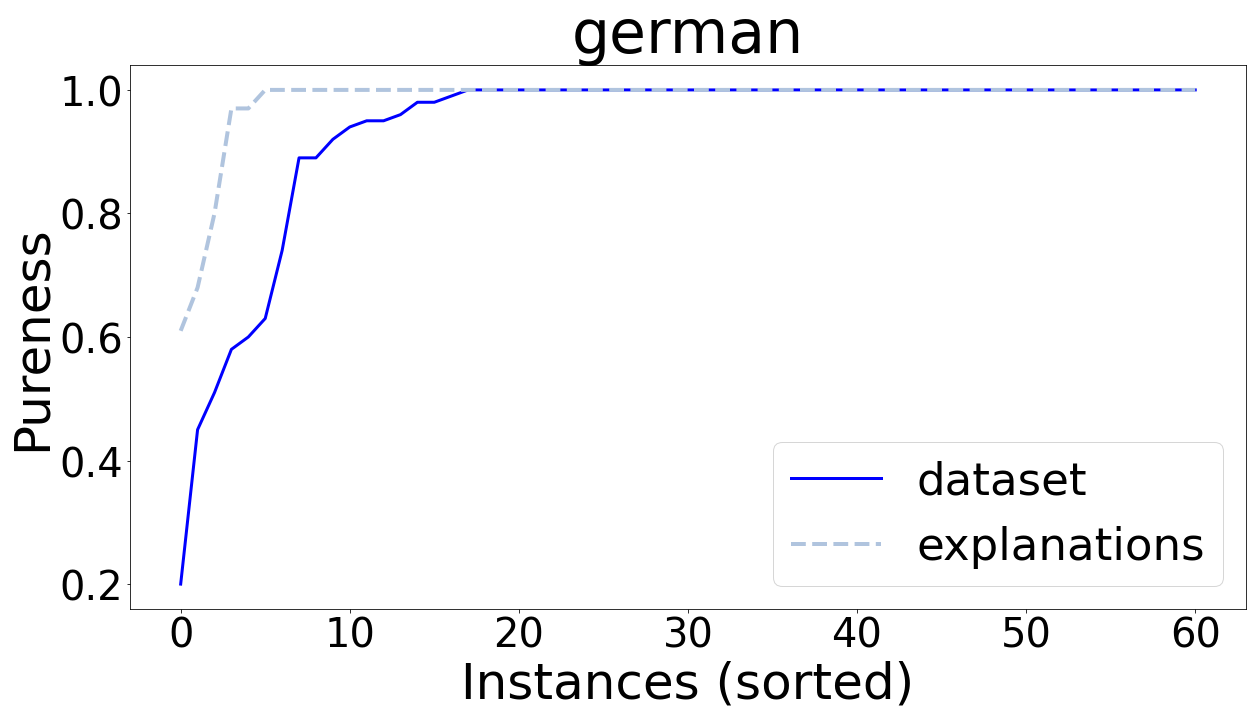}
\end{subfigure}\hspace*{\fill}
\begin{subfigure}{0.48\textwidth}
\includegraphics[width=\linewidth]{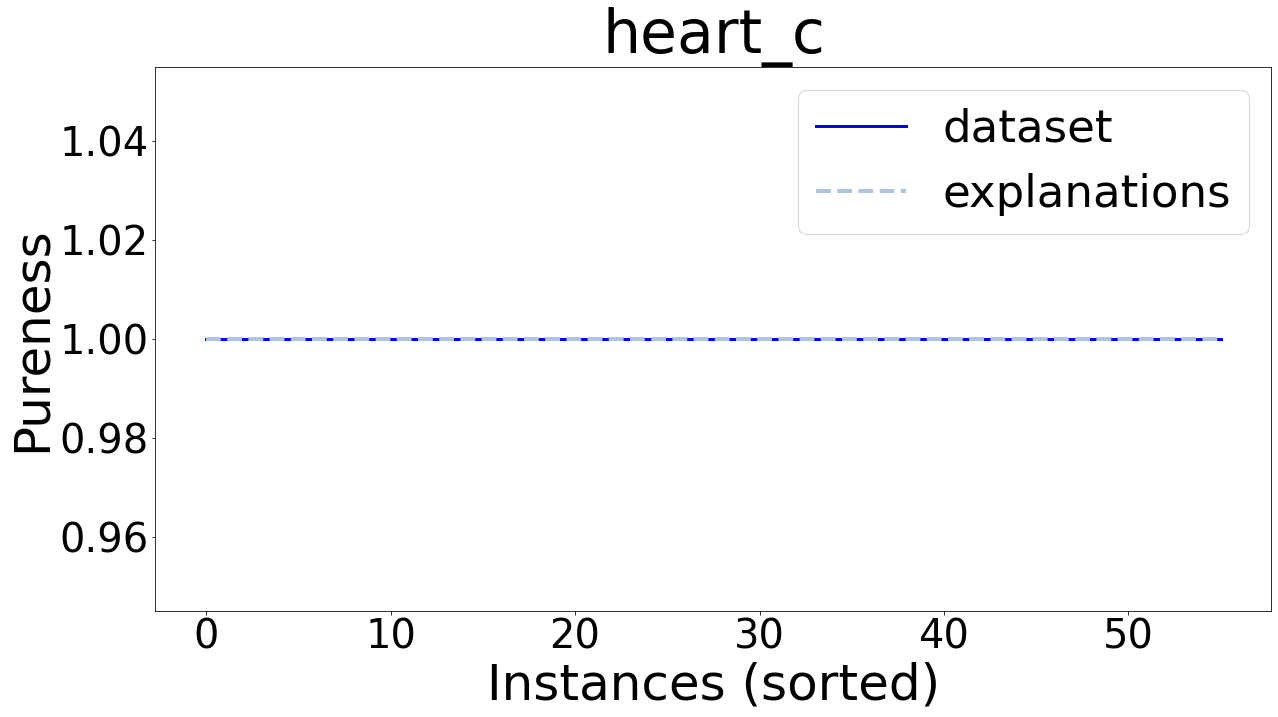}
\end{subfigure}
\medskip
\begin{subfigure}{0.48\textwidth}
\includegraphics[width=\linewidth]{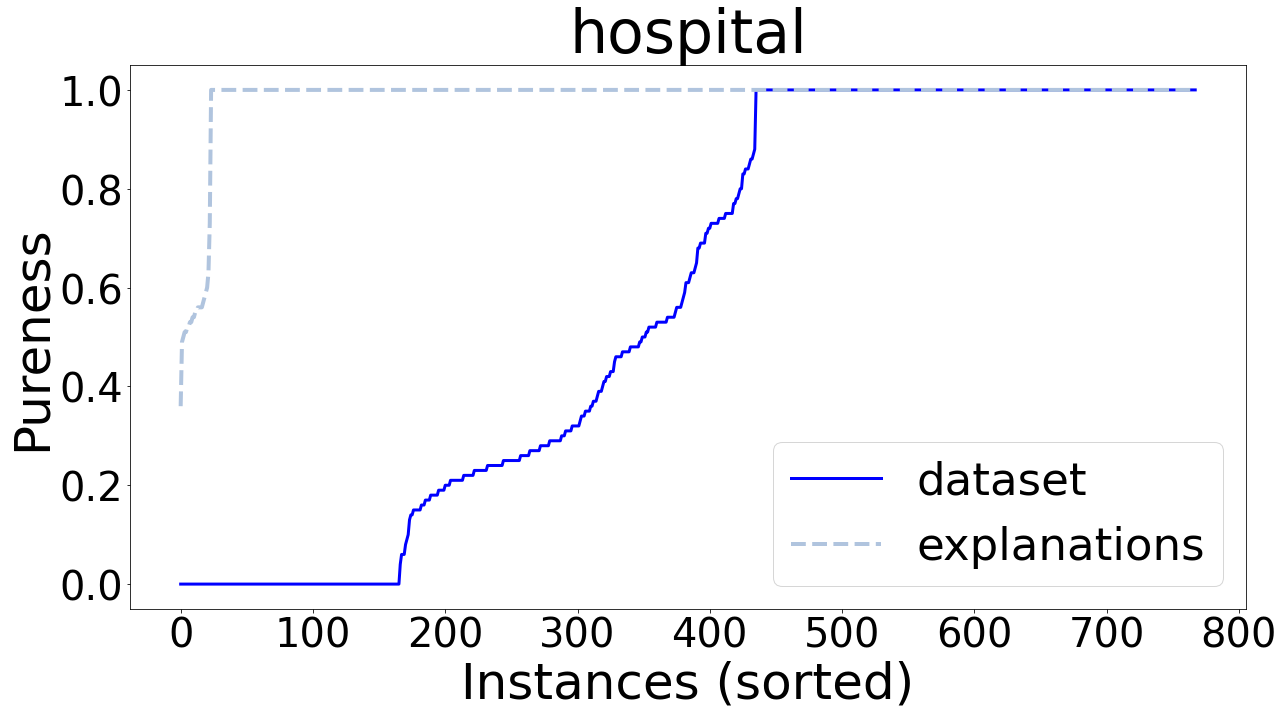}
\end{subfigure}
\begin{subfigure}{0.48\textwidth}
\includegraphics[width=\linewidth]{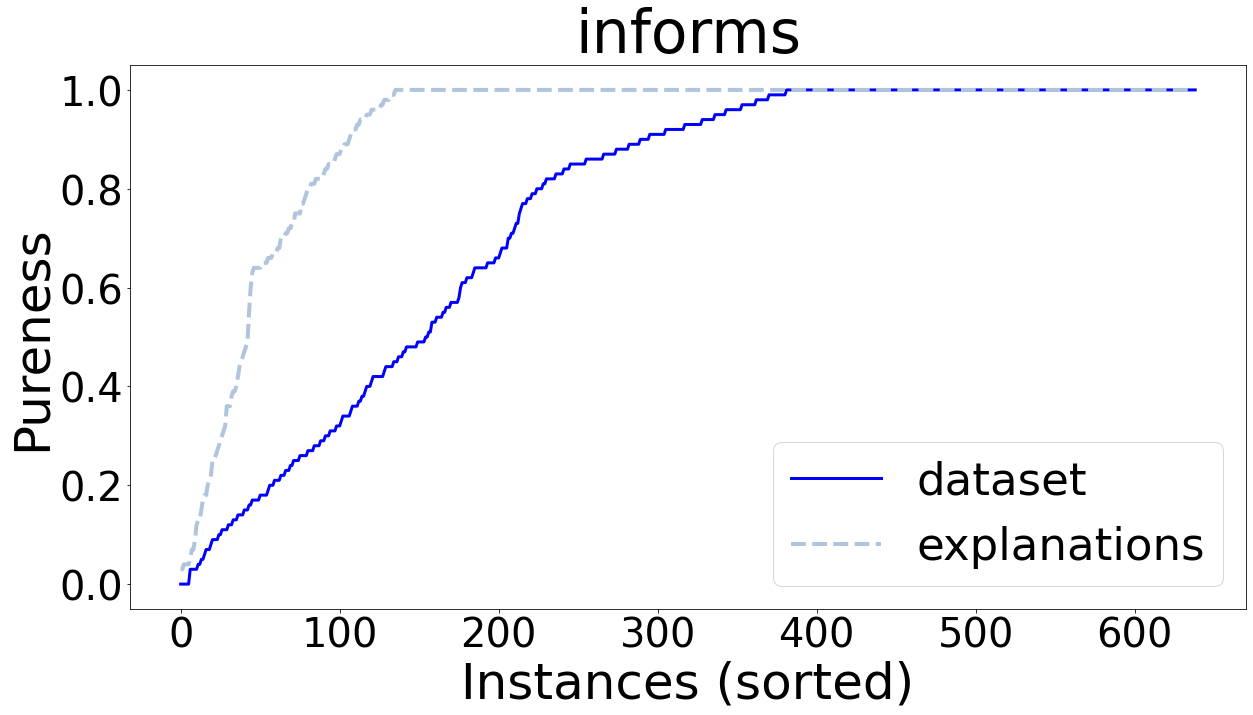}
\end{subfigure}\hspace*{\fill}
\caption{Comparison of the \emph{counterfactual validity} between Mondrian and GRASP implementation.} \label{pureness_mondrian}
\end{figure}

We compare CF-K with an alternative strategy: making the whole dataset $k$-anonymous, and taking the counterfactual explanations out of this  anonymized dataset. This differs from our strategy where we directly make the counterfactual instances and explanations $k$-anonymous.
We use an open source implementation of Mondrian\footnote{\url{https://github.com/danielegiampaoli/Mondrian_K-anonymization}} to compare CF-K with. Mondrian is a top-down greedy data anonymization algorithm that has been shown to be one of the best performers~\citep{ayala2014systematic,lefevre2006mondrian}.
For all instances in the test set (max 1,000) with an unfavorable outcome, we  compare the $k$-anonymous counterfactual explanation generated by CF-K with the $k$-anonymous counterfactual explanation based on an instance selected from the anonymized (by Mondrian) test set.

We see in Figure~\ref{ncp_mondrian} that for all datasets CF-K succeeds in achieving a better (and thus lower) average NCP than the Mondrian implementation on the whole dataset. This is in line with our hypothesis that only $k$-anonymizing the counterfactual instances can result in lower information loss, as unused training instances do not need to be used in the calculations for the best clustering.

In Figure~\ref{pureness_mondrian}, we see that the average counterfactual validity (measured by pureness) in all datasets is higher when using $k$-anonymous explanations than when using an explanation from a $k$-anonymous dataset (except for the \emph{Heart} dataset, where the average counterfactual validity is 100\% in both implementations). Counterfactual validity can only be calculated on an explanation, and not on a dataset, so methods to make the dataset $k$-anonymous can not optimize for this metric. Therefore, our methodology to make the explanations $k$-anonymous, was needed to take this metric into account. 

\subsection{Interplay between the metrics} \label{sec:trade-off}

\begin{figure}[h] 
\begin{subfigure}{0.48\textwidth}
\includegraphics[width=\linewidth]{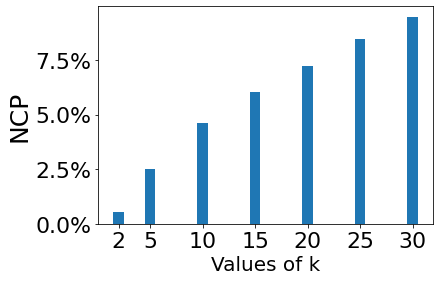}

\end{subfigure}\hspace*{\fill}
\begin{subfigure}{0.48\textwidth}
\includegraphics[width=\linewidth]{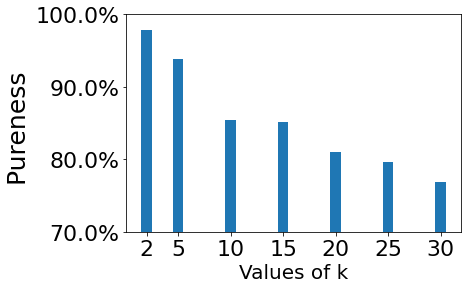}
\end{subfigure}
\caption{Trade-off between the metrics (tested on \emph{CMC} dataset with $\alpha = 40$ and \emph{iterations = 5}).} \label{trade-off}
\end{figure}

We test multiple parameter values for $k$ to establish its effect on the NCP. In Figure~\ref{trade-off}, we demonstrate the experiments on the \emph{CMC} dataset (the other datasets show the same patterns). We see that increasing $k$, which increases the privacy guarantees for each individual, deteriorates the other two metrics. The Normalized Certainty Penalty, which measures how much information value we lose by making the data $k$-anonymous, increases when we increase the value of $k$. This makes sense as the data quality degrades more when we add more privacy guarantees and thus require more instances to be identical. Furthermore, the average pureness, and thus the counterfactual validity, also decreases. The trade-off between privacy (measured by $k$) and information loss (measured by NCP) has been confirmed by the literature~\citep{ayala2014systematic,sumana2010anonymity}, but we are the first to show this trade-off between $k$ and counterfactual validity (measured by pureness).

\subsection{Does this have fairness implications?} \label{sec:fairness}

\begin{figure}[htb!] 
\begin{subfigure}{0.48\textwidth}
\includegraphics[width=\linewidth]{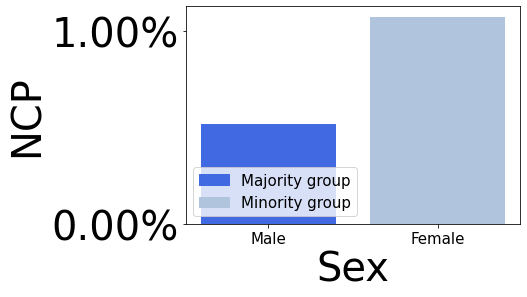}
\caption{Adult Income} 
\end{subfigure}\hspace*{\fill}
\begin{subfigure}{0.48\textwidth}
\includegraphics[width=\linewidth]{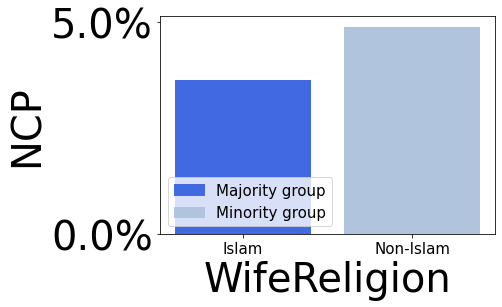}
\caption{CMC}
\end{subfigure}
\medskip
\begin{subfigure}{0.48\textwidth}
\includegraphics[width=\linewidth]{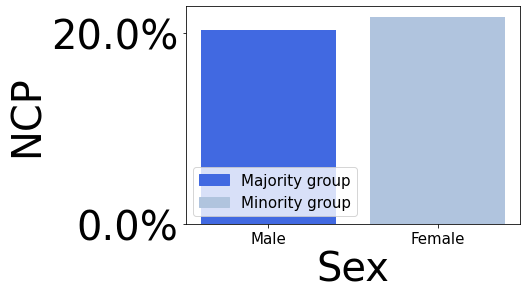}
\end{subfigure}\hspace*{\fill}
\begin{subfigure}{0.48\textwidth}
\includegraphics[width=\linewidth]{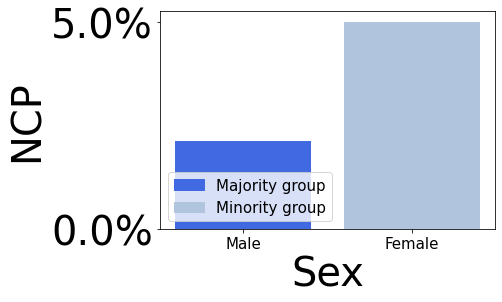}
\caption{Heart} 
\end{subfigure}
\medskip
\begin{subfigure}{0.48\textwidth}
\includegraphics[width=\linewidth]{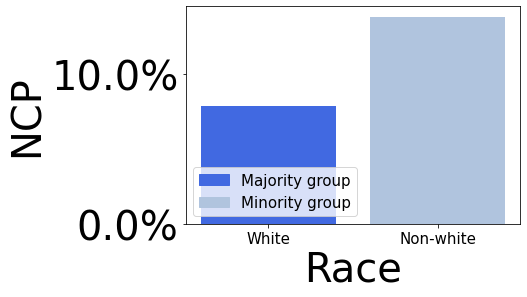}
\caption{Informs}\label{fig:f}
\end{subfigure}
\hspace{4mm}
\begin{subfigure}{0.48\textwidth}
\includegraphics[width=\linewidth]{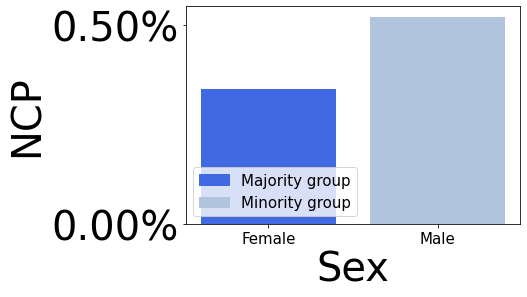}
\caption{Hospital} 
\end{subfigure}\hspace*{\fill}
\caption{Comparison of the average NCP between the majority and minority group} \label{ncp_fairness}
\end{figure}
A minority group is defined as a group whose characteristics such as race, religion, gender, ... etc. are fewer in numbers than the main group of that classification. Nowadays, it is often used to refer to people that experience a relative disadvantage based on their group membership~\citep{healey2019race}. We define the minority and majority group for each dataset based on the sensitive attribute, mentioned in Table~\ref{used_datasets}.
We see in Figure~\ref{trade-off} that when we make the explanations more private (increase $k$), the explanation quality decreases and they become less useful. Unfortunately, this effect is larger for minority groups which can lead to potential issues regarding fairness.
As can be seen in Figure~\ref{ncp_fairness}, in every examined dataset, the average NCP is higher for the minority group. For the average counterfactual validity, we found no difference between both groups. 
Figure~\ref{ncp_fairness} shows that the quality of their explanations has to be reduced more to achieve the same level of privacy. This can be explained by the fact that they often have more unique quasi-identifiers, as there are less people that share their public characteristics (definition of a minority group), so their quasi-identifiers have to be generalised more to be anonymous. When explanations are used in high-stakes settings, it is undesirable that minority groups are offered lower quality explanations, but also that there is a risk of leaking their private information~\citep{patel2020model}.
These results show that different ethical objectives can work against each other and that one has to make sure that minority groups are not adversely affected in unexpected ways.

\section{Conclusion}

Transparency in machine learning has become an important topic, yet there is little research on the potential risks to user privacy that this poses~\citep{patel2020model}. Although research has shown that offering model explanations may come at the cost of user privacy~\citep{sokol2019counterfactual}, none of the currently offered model explanation technologies offer any privacy guarantees. Once such explanation systems are deployed on high-stakes data, like financial transactions or patient health records, a formal investigation of privacy risks is necessary.  
In this research, we introduce the \emph{explanation linkage attack}, constituting the privacy risk that some counterfactual explanation techniques pose to the privacy of data subjects, as adversaries can infer their private attributes. We are the first to apply $k$-anonymity on counterfactual explanations instead of on the complete dataset and show that applying $k$-anonymity only on the counterfactual explanations can achieve lower \emph{information loss} and higher \emph{counterfactual validity}. Furthermore, we see that if we increase the privacy constraints, the quality of the explanations becomes worse which shows the trade-off between \emph{privacy} and \emph{transparency}.

Other researchers~\citep{patel2020model,shokri2019privacy} have stated that assessing the privacy/explainability trade-off for minority groups is a promising avenue for future exploration, which is what we explored in Section~\ref{sec:fairness}. We noticed that the average information loss tends to be higher for minority groups, and this difference increases with the level of privacy, hereby introducing a new element of unfairness.

A debate on explanation quality could also be a promising avenue for future research. For $k$-anonymous counterfactual explanations that have a pureness of 100\%, generalized quasi-identifiers might actually be an advantage instead of a drawback. Think about the following scenario:
Would you prefer the explanation \emph{`If you would be a teacher and earned \$10K more, then you would have received the loan'} or the explanation \emph{`If you would be a teacher \emph{or a nurse} and earned \$10K more, then you would have received the loan'}, if both explanations are valid? While generalizing instances in a dataset means less information value, this trade-off is less clear in counterfactual explanations: generalizing them might give you more options to achieve the required target outcome and thus be \emph{more} valuable. However, this is only the case when the counterfactual explanations are entirely \emph{valid} and have a pureness of 100\%. As a discussion on explanation quality was not the goal of this study, we leave this for future research.

\section*{Funding}
This research was funded by Research Foundation—Flanders grant number 11N7723N.

\appendix

\section{CF-K: implementation details}
\subsection{Algorithm description}
Our algorithm starts from a counterfactual explanation that is given to one of the instances in the test set with an unfavorable prediction outcome. The counterfactual instance that this explanation is based on is an actual instance in the training set, and we want it to be unidentifiable from at least $k-1$ other instances in the training set. This is the case when at least $k-1$ other instances in the training set have the same values for the quasi-identifiers (these are the attributes that we assume to be publicly known).

\paragraph{Phase 1: Construct greedy randomized solution}
In this phase, we construct a feasible solution. We  first check for the current counterfactual instance if its values of quasi-identifiers are present for $k$ individuals in the training set. In this case, a solution is found, the quality of the solution is calculated and the algorithm moves to the next phase.
If this is not the case, we generate a list of size $\alpha$ by selecting the closest neighbors of the counterfactual instance in the training set with the required prediction outcome.  Then, we \emph{randomly} select a neighbour from this list and create a new \emph{generalized instance} out of this neighbor and the counterfactual instance. The fact that we randomly select a neighbor out of this list and not just select the closest neighbour makes up the probabilistic component of GRASP. We create this \emph{generalized instance} by generalizing the values of the quasi-identifiers so that the generalized instance includes both the values of the quasi-identifiers of the counterfactual instance as those of the neighbor. This happens as in Figure~\ref{gen_instance2}.
We check again whether this generalised instance satisfies $k$-anonymity. If this is the case, a solution is found, the quality of this solution is calculated and the algorithm moves to the next phase. If this is not the case, this loop is repeated until the generalised instance satisfies $k$-anonymity.

\paragraph{Phase 2: Local search}
The local search algorithm iteratively tries to replace the current solution by a better solution in the neighbourhood. The algorithm terminates when no better solution is found.
The neighborhood is defined by checking for every quasi-identifier in the current solution whether slightly changing it, is a feasible solution (satisfies $k$-anonymity) and improves the solution quality. A slight change in this case is adding a value (if the quasi-identifier is a single value) or removing a value from the list (if the quasi-identifier is already a generalized list).

\paragraph{GRASP}

\begin{algorithm}
\caption{GRASP}
\begin{algorithmic}
\For{$i = 1,...,MaxIter$} 
\State $Solution \leftarrow ConstructGreedyRandomizedSolution(Input)$;
\State $Solution \leftarrow LocalSearch(Solution)$;
\State $BestSolution \leftarrow UpdateSolution(Solution,BestSolution)$;
\EndFor \\
\Return $BestSolution$;
\end{algorithmic}
\end{algorithm}

We iterate these two phases for a specified number of iterations. After each iteration, we check if the new solution is better than the current best solution and if this is the case, we update the current best solution.
After the specified number of iterations, the algorithm terminates and the current best solution is returned. For our experiments, we used $k=10$, $\alpha = 20$ and maxiter $= 3$.

\subsection{Choice of parameters}
\paragraph{Parameter $\alpha$}
The parameter $\alpha$ is a measure of the randomness of the algorithm, as it determines the number of closest neighbors from which we randomly select one. We see that increasing $\alpha$ will slightly lower the NCP until a certain level, after which it starts to increase again (too much randomness). Overall, we identified the optimal value of $\alpha$ to be around 20 for $k=10$, but the results remain robust for other values of $\alpha$ as well.
$\alpha$ does not have a significant effect on time.

\paragraph{Number of iterations}
We see that the NCP decreases when we increase the
number of iterations. This effect becomes smaller when the number of iterations
is higher. We also see that the execution time increases with the number of
iterations so a trade-off has to be made between solution quality and execution
time in determining the optimal number of iterations. In our experiments, we use 3 iterations.


\bibliographystyle{unsrtnat}

\end{document}